\pdfoutput=1

\documentclass[11pt]{article}

\usepackage{EMNLP2023}

\usepackage{times}
\usepackage{latexsym}
\usepackage{xcolor}

\usepackage[T1]{fontenc}

\usepackage[utf8]{inputenc}

\usepackage{microtype}
\usepackage{enumitem}
\usepackage{fancyhdr}
\usepackage{inconsolata}
\usepackage{pgfplots}
\usepackage{xcolor}

\newcommand\blfootnote[1]{%
  \begingroup
  \renewcommand\thefootnote{}\footnote{#1}%
  \addtocounter{footnote}{-1}%
  \endgroup
}

\title{nlpBDpatriots at BLP-2023 Task 1: A Two-Step Classification for Violence Inciting Text Detection in Bangla}

\author{Md Nishat Raihan\textsuperscript{*}, Dhiman Goswami\textsuperscript{*}, Sadiya Sayara Chowdhury Puspo\textsuperscript{*}, \\ \textbf{Marcos Zampieri} \\
        George Mason University \\
         \texttt{\{dgoswam|mraihan2|spuspo|mzampier\}@gmu.edu} \\
        }

\begin{document}
\maketitle
\begin{abstract}

In this paper, we discuss the nlpBDpatriots entry to the shared task on Violence Inciting Text Detection (VITD) organized as part of the first workshop on Bangla Language Processing (BLP) co-located with EMNLP. The aim of this task is to identify and classify the violent threats, that provoke further unlawful violent acts. Our best-performing approach for the task is two-step classification using back translation and multilinguality which ranked $6^{th}$ out of 27 teams with a macro F1 score of 0.74.
\blfootnote{*These three authors contributed equally to this work.} 
\end{abstract}

\section{Introduction}

In an era dominated by social media platforms such as Facebook, Instagram, and TikTok, billions of individuals have found themselves connected like never before, enabling them to swiftly share their thoughts and viewpoints. The growth of social networks provides people all over the world with unprecedented levels of connectedness and enriched communication. However, social media posts often abound with comments containing varying degrees of violence, whether expressed overtly or covertly \cite{kumar2018benchmarking,kumar-etal-2020-evaluating}. 
To combat this worrisome trend, social media platforms established community guidelines and standards that users are expected to adhere to.\footnote{\url{https://transparency.fb.com/policies/community-standards/hate-speech}}\textsuperscript{,}\footnote{\url{https://help.twitter.com/en/rules-and-policies/hateful-conduct-policy}}.
Violations of these rules may result in the removal of offensive content or even the suspension of user accounts. Given the vast amount of user-generated content on these platforms, manually scrutinizing and filtering potential violence is a very challenging task. This moderation approach is limited by moderators' capacity to keep pace, comprehend evolving slang and language nuances, and navigate the complexity of multilingual content \cite{das2022hate}. To address this issue, several social media platforms turn to AI and NLP models capable of detecting inappropriate content across a range of categories such as aggression and violence, hate speech, and general offensive language \cite{zia2022improving,weerasooriya2023vicarious}.%



The shared task on Violence Inciting Text Detection (VITD) \cite{SahaAndJunaed} aims to categorize and discern various forms of communal violence, aiming to shed light on mitigating this complex phenomenon for the  Bangla speakers. For this task, we carry out various experiments presented in this paper. We employ various models and data augmentation techniques for violent text identification in Bangla.


\section{Related Work}

\paragraph{Violence Identification in Bangla} Several works have been done on building datasets similar to this task and training models on those data. Such datasets include the works of \cite{remon2022bengali, das2022hate}, which mostly gather data by social media mining. However, most of the datasets are comparatively small in size. One of the larger datasets is prepared by \citet{romim2022bd}, which consists of 30,000 user comments from YouTube and Facebook, annotated using crowdsourcing. 

While most works focus primarily on the datasets, they also present some experimental analysis. \citet{das2022hate} evaluates transformer-based models like m-BERT, XLM-RoBERTa, IndicBERT, and MuRIL. XLM-RoBERTa excels with ample training and MuRIL performs well in joint training, while m-BERT and IndicBERT show proficiency in zero-shot scenarios. However, the most notable work here is done by \citet{jahan-etal-2022-banglahatebert} who introduces BanglaHateBERT, a re-trained BERT model for abusive language detection in Bangla. It is trained on a large-scale Bangla offensive, abusive, and hateful corpus. The authors collect and annotate a balanced Bangla hate speech dataset and use it to pretrain BanglaBERT. The proposed model, BanglaHateBERT, outperforms other BERT models and CNN-based models in detecting hate speech on benchmark datasets.

\paragraph{Related Shared Tasks} \citet{zampieri2019semeval,zampieri-etal-2020-semeval} organized OffensEval, a series of shared tasks identifying and categorizing offensive language in tweets organized at SemEval 2019 and 2020. At OffensEval, participants trained a variety of models ranging from machine learning to deep learning approaches. While BERT and other transformed dominated the leaderboard in 2020, systems' performance in 2019 was more varied with traditional ML classifiers and ensemble-based approaches achieving competition performance along with deep learning approaches. Another shared task, MEX-A3T track at IberLEF 2019 \cite{inproceedings_spanish}, focused on author profiling and aggressiveness detection in Mexican Spanish tweets. 
Additionally, \citet{modha2021overview} presents an overview of the HASOC track at FIRE 2021 for hate speech and offensive content detection in English, Hindi, and Marathi, where the highest accuracy is achieved on the Marathi dataset. 


\section{Dataset}

The VITD shared task \cite{data} provides the participants with a Bangla dataset including 2700 instances for training and 1330 instances for development. The blind test set contains 2016 instances. The dataset \cite{SahaAndJunaed}  has been annotated using three labels: Non-Violence, Direct-Violence, and Passive-Violence. 
This three-class annotated dataset differs from similar datasets where a binary annotation is used \cite{romim2022bd,mridha2021boost}. The data distribution per label is shown in Table \ref{tab:label_distribution}.

\begin{table}[!ht]
\centering
\begin{tabular}{lccc}
\hline
\textbf{Label} & \textbf{Train} & \textbf{Dev} & \textbf{Test} \\
\hline
Non-Violence & 51\% & 54\% & 54\% \\
Passive-Violence & 34\% & 31\% & 36\% \\
Direct-Violence & 15\% & 15\% & 10\% \\
\hline
\end{tabular}
\caption{Label-wise data distribution across training, development, and test datasets.}
\label{tab:label_distribution}
\end{table}


\begin{figure*} [!h]
  \centering
  \includegraphics[width=\textwidth]{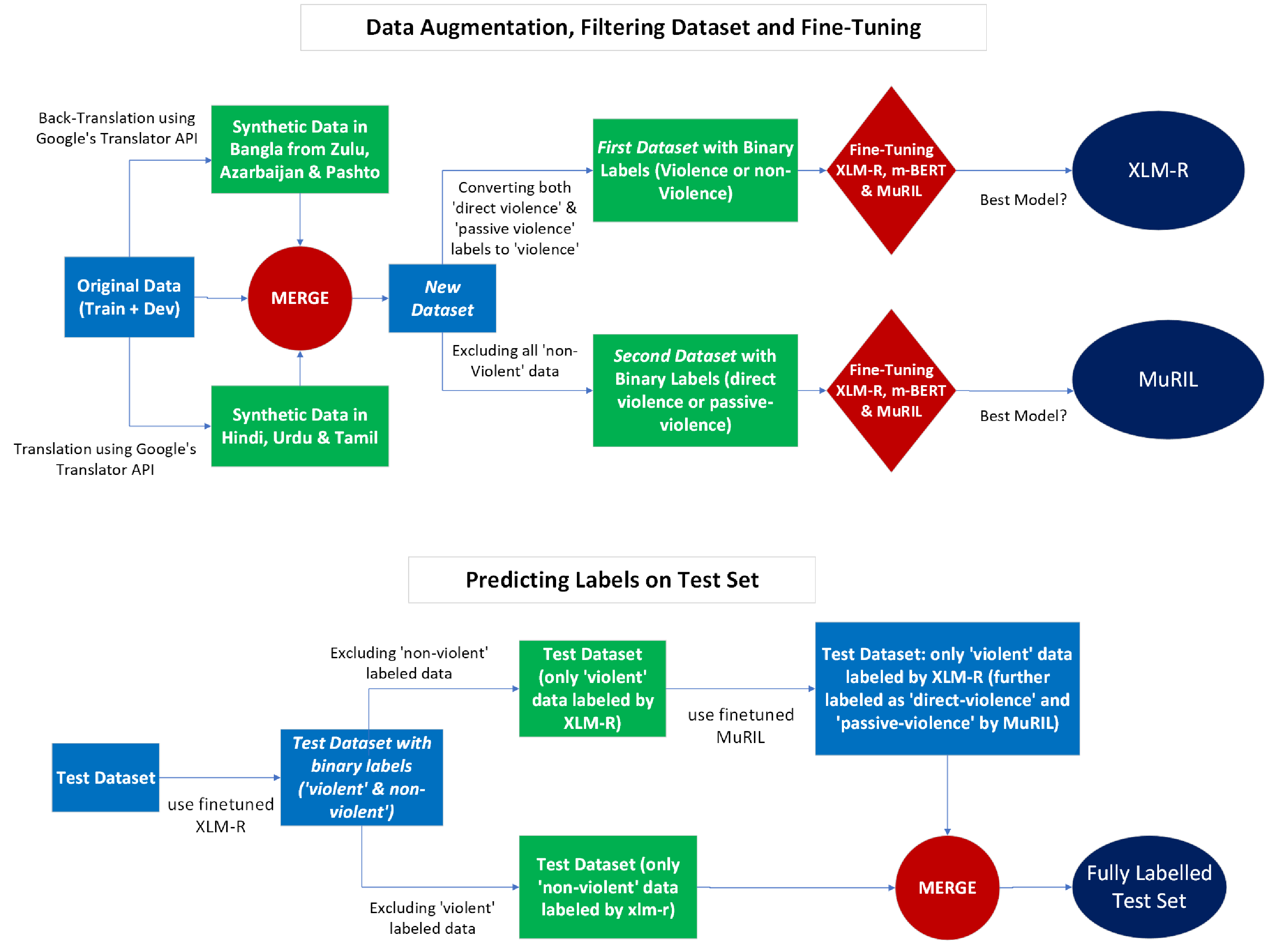}
  \caption{Two-step Classification with Data Augmentation}
  \label{fig:flowchart}
\end{figure*}

\section{Methodologies}


\subsection{Models}

\paragraph{Statistical ML Classifiers} In our experiments, we use statistical machine learning models like Logistic Regression and Support Vector Machine using TF-IDF vectors. 

\paragraph{Transformers} We test multiple transformer models pre-trained on Bangla. Our initial experiments include Bangla-BERT \cite{kowsher2022bangla} which is only pre-trained on Bangla corpus. We finetune the model on the train set and evaluate it on the dev set with empirical hyperparameter tuning. 
We then use multilingual transformer models like multilingual-BERT \cite{devlin2019bert} and xlm-roBERTa \cite{conneau2020unsupervised}, which are pre-trained on 104 and 100 different languages respectively, including Bangla. We also do the same hyperparameter tuning with both models. 
Lastly, we use MuRIL \cite{khanuja2021muril}, another transformer pre-trained in 17 Indian languages including Bangla. 

\paragraph{Task Fine-tuned Models} We use BanglaHateBERT \cite{jahan-etal-2022-banglahatebert} as a task fine-tuned model which is developed on existing pre-trained BanglaBERT \cite{kowsher2022bangla} model and retrained with 1.5 million offensive posts.

\paragraph{Prompting} We prompt gpt-3.5-turbo model \cite{openai2023gpt35turbo} from OpenAI for this classification task. We use the API to prompt the model, while providing a few examples for each label and ask the model to label the dev and test set. 

\subsection{Data Augmentation} \label{data_aug}

Given the relatively small size of the VITD dataset, we implement a few data augmentation strategies to expand its size. First, we use Google's Translator API \cite{google2021translateapi} to translate the train and dev set to 3 other languages that are very similar to Bangla (Hindi, Urdu, and Tamil). Bangla, Hindi, Urdu belong to Indo-Aryan language branch and Tamil from Dravidian language brach, though, all of these languages have cultural interaction in south-east asian region. The native speakers of these languages live in closer geographic proximity. Moreover, these languages have similar morphosyntactic features. So, translating Bangla text to those languages do not hamper structural and grammatical integrity of the sentences. Therefore, we combine these new synthetic datasets with the original train dataset and finetune the multilingual transformer models on them. 

The second approach to augment the dataset is back translation. We again use the Translator API to translate the original train and dev set to a few different low-resource languages like Zulu, Pashto, and Azarbijani as the intermediary language for back translation, in order to add more context. Zulu is from Niger-Congo, Pashto is Indo-Iranian and Azabijani is from Turkic language family. As these languages does not have any cultural interaction with Bangla, back translating from these languages will make three additional version of same sentences with versatility. Then we combine these data with the original dataset. We observe that xlm-roBERTa produces a better macro F1 than the first approach, but still the same as it was on the original data, 0.73. 

\subsection{Two-step Classification with Data Augmentation}

Finally, we combine the two dataset augmentation techniques discussed previously. After combining the synthetic data with the original train set, we have a \textit{New Dataset} that is 7 times the size of the original train set. We generate two different datasets using this \textit{New Dataset}. For the \textit{First Dataset}, we convert all the labels in the \textit{New Dataset} to either Violent (1) or non-Violent (0). And for the \textit{Second Dataset}, we only keep the violent data (both Direct and Passive) from the \textit{New Dataset}. 

We finetune mBERT, MuRIL and xlm-roBERTa on both binary labeled \textit{First Datatset} and \textit{Second Dataset} and save their model weights. xlm-roBERTa outperforms the other two when fine-tuned the \textit{First Dataset} and \textit{MuRIL} outperforms the other two when fine-tuned on the \textit{Second Dataset}. For the test set, we first use the finetuned xlm-roBERTa to label the whole dataset as either violent or non-violent data. We then separate all the data from the test set that are labeled as 'violent' by the finetuned xlm-roBERTa model and use the fine-tuned MuRIL model to predict the 'active violence' and 'passive violence' labels. Finally, we merge this with all the 'non-violent' labeled datasets from the first step. Thus, we get all the predicted labels for the test set using 2-step classification by two fine-tuned models. The whole procedure is demonstrated in Figure \ref{fig:flowchart}. 

\section{Results and Analysis}

\subsection{Results}

At the start of the shared task, three baseline macro F1 scores have been provided by the organizers. For BanglaBERT, XLM-R and mBERT, the provided baselines are 0.79, 0.72, and 0.68 respectively. The results of our experiments are shown in Table \ref{Results}.

\begin{table} [!h]
\centering
\scalebox{.90}{
\begin{tabular}{lccc}
\hline
\textbf{Models} & \textbf{Dev} & \textbf{Test} \\
\hline
Logistic Regression & 0.55 & 0.56 \\
Support Vector Machine & 0.61 & 0.63 \\
\hline
BanglaBERT & 0.66 & 0.67 \\
mBERT & 0.71 & 0.67 \\
MuRIL & 0.81 & 0.72 \\
XLM-R & 0.79 & 0.73 \\
\hline
BanglaHateBERT & 0.59 & 0.60 \\
\hline
GPT 3.5 Turbo & 0.46 & 0.43 \\
\hline
XLM-R (Self-transfer Learning) & 0.79 & 0.72 \\
XLM-R (Multilinguality) & 0.78 & 0.72 \\
XLM-R (Back Translation) & 0.77 & 0.73 \\
XLM-R, MuRIL (Two-step) & \textbf{0.84} & \textbf{0.74} \\
\hline
\end{tabular}
}
\caption{Dev and test macro F-1 score for all evaluated models and procedures.}
\label{Results}
\end{table}

\noindent Among the statistical machine learning models, we use logistic regression and support vector machine. For logistic regression, we achieve a macro F1 score of 0.56 and for the support vector machine the F1 is 0.63. For transformer-based models, we use BanglaBERT, mBERT, MuRIL and XLM-R where we get the best F1 score of 0.73 by XLM-R. Task fine-tuned model BanglaHateBERT scores 0.60 macro F1.

A few shot learning procedure is used by using GPT3.5 Turbo. We give a few instances of each label as prompt and got 0.43 F1 which is significantly lower than our other attempted approaches. This is because GPT3.5 is still not enough efficient for any downstream classification problem in Bangla like this shared task.

We also perform some customization in our approach instead of directly using the existing models. We use transfer learning. Instead of using the basic idea of transfer learning by fine-tuning a model with a larger dataset of the same label, we translate the train set to English with Google Translator API and used XLM-R on that data. Then we use that finetune model and perform the same procedure over the actual Bangla train set. We refer this procedure as \textit{self-transfer learning} and the F1 score from this procedure is 0.72.

Introducing multilinguality to many downstream tasks proves to be effective. So we also opt for this procedure by translating the train data using Google Translator API to Hindi, Urdu, and Tamil as they are grammatically less diverse and vocabulary is close in contact among the native speakers of these languages. That is how we make the size of our train set three times higher than the original one and got a 0.72 F1 score.

On the other hand, we use Zulu, Azerbaijan, and Pashto - 3 very diverse languages from Bangla for back translation. So, we also get the size of our train set three times higher than the original Bangla one with significantly different translations for each instance. And we get a 0.73 F1 score for that. 


Moreover, we use a two-step classification with the data achieved by multilinguality and back translation. Along with these data, we also merge our original Bangla train set. Then, we perform two separate streams of classification. At first, instead of direct and passive violence, we convert them as violence and finetune by XLM-R, mBERT, and MuRIL to classify violence and non-violence where XLM-R performs the best. Then we use the same procedure with the same models to classify direct and passive violence from the merged labels of violence where MuRIL performs the best. Following this procedure, we achieve our best macro F1 score of 0.74 for this shared task. 

\subsection{Analysis}

\begin{figure*}[!h]
  \centering
  \includegraphics[width=.75\linewidth]{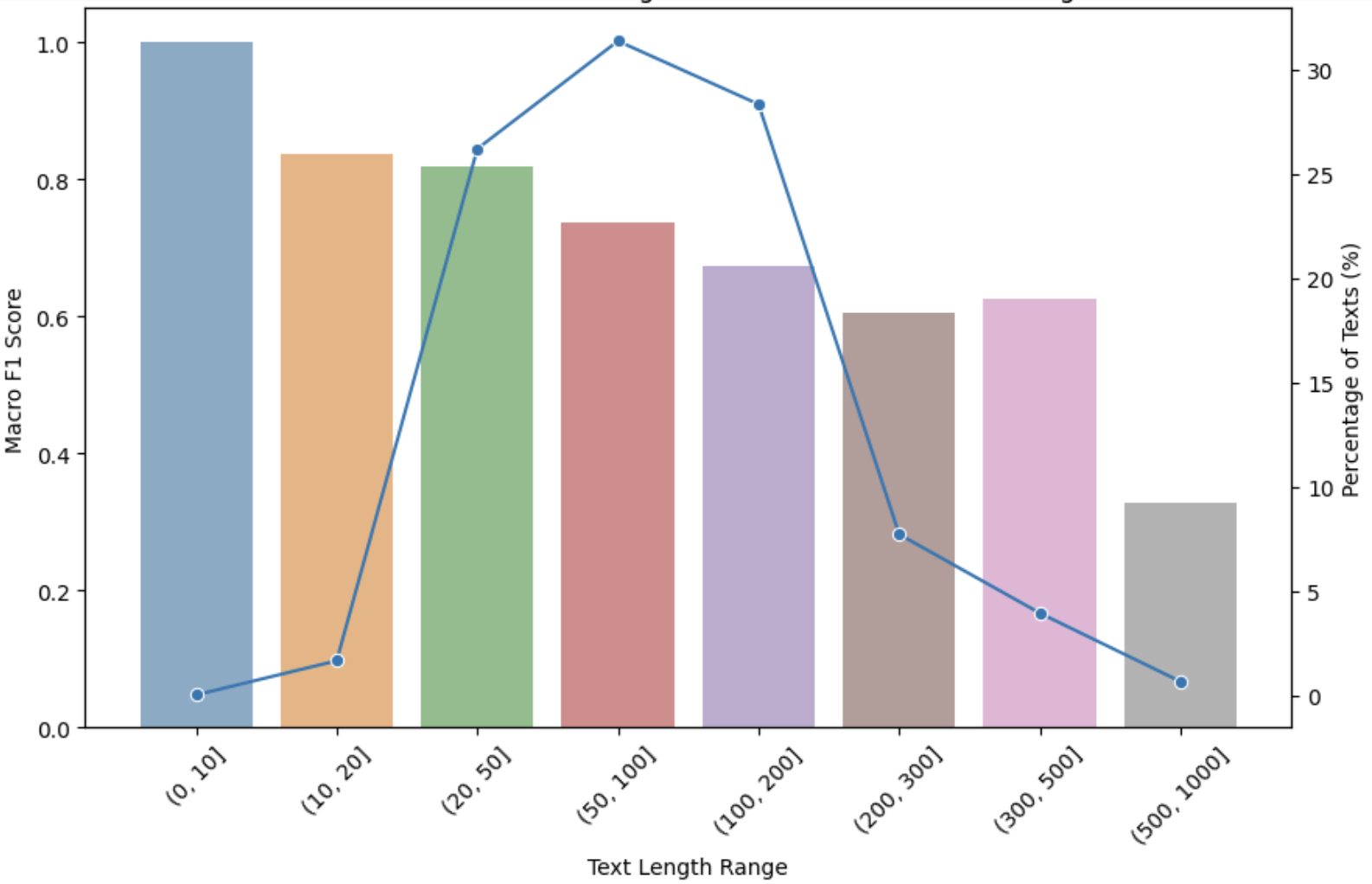}
  \caption{Performance analysis based on text length.}
  \label{fig:Performance Analysis}
\end{figure*}

In terms of text length, the model attains a perfect macro F1 score of 1.000 for texts of 10 words or fewer but struggles with longer texts, evidenced by a macro F1 of only 0.329 for texts of 500-1000 words (Figure \ref{fig:Performance Analysis}, Table \ref{tab:Text length analysis}). Though, it maintains respectable F1 scores for text lengths commonly encountered in the dataset, future work should focus on enhancing F1 score for texts with direct violence content. 

\begin{table}[!ht]
    \centering
    \begin{tabular}{lcccc}
        \hline
        Text Length & Macro F1 & Count & Percentage \\
        \hline
        (0, 10]     & 1.000 & 1   & 0.050 \\
        (10, 20]    & 0.836 & 34  & 1.687 \\
        (20, 50]    & 0.820 & 528 & 26.190 \\
        (50, 100]   & 0.736 & 632 & 31.349 \\
        (100, 200]  & 0.673 & 571 & 28.323 \\
        (200, 300]  & 0.606 & 156 & 7.738 \\
        (300, 500]  & 0.627 & 80  & 3.968 \\
        (500, 1000] & 0.329 & 14  & 0.694 \\
        \hline
    \end{tabular}
    \caption{Performance analysis based on text length.}
    \label{tab:Text length analysis}
\end{table}

\begin{figure}[!ht]
  \centering
  \includegraphics[width=\linewidth]{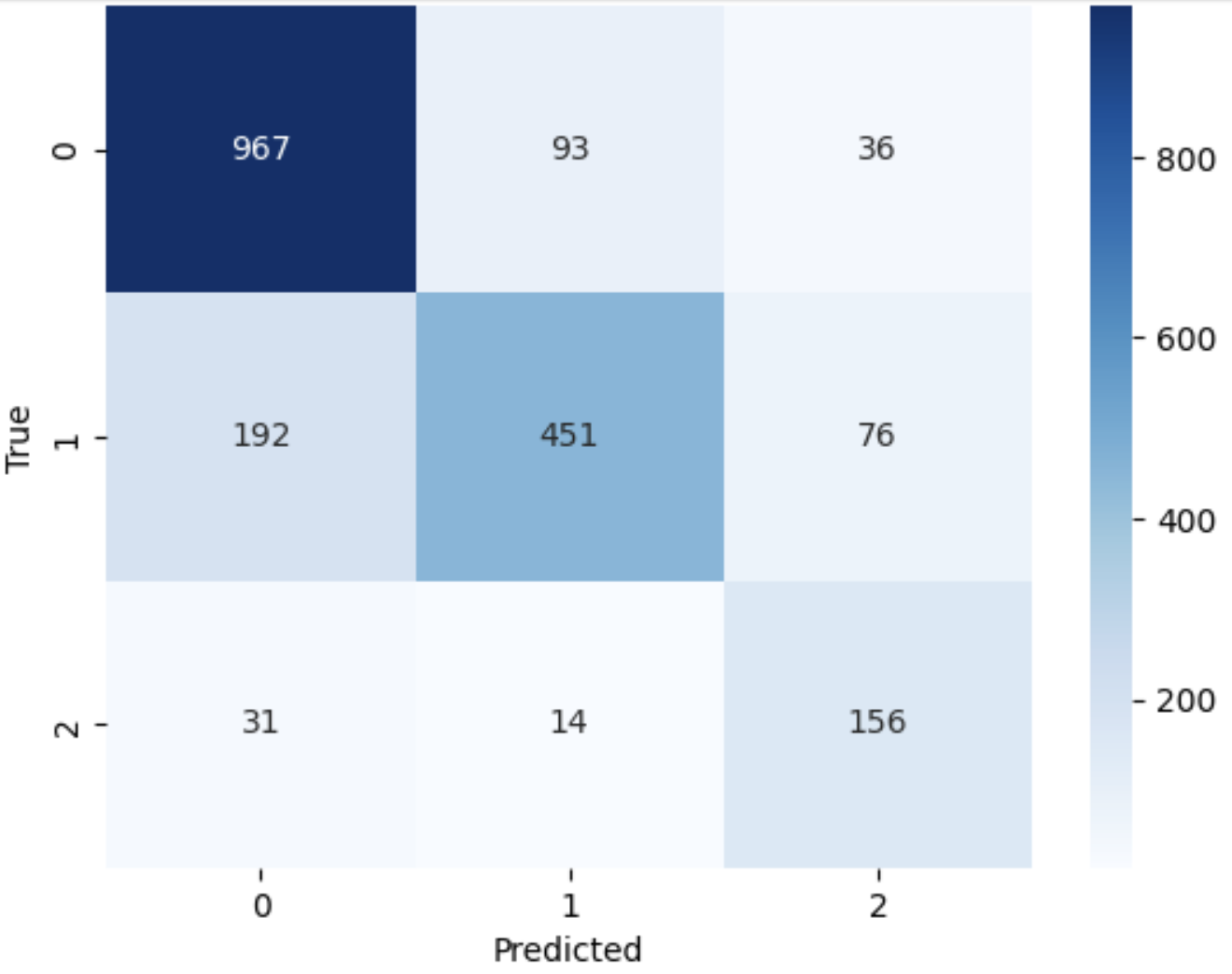}
  \caption{Confusion Matrix}
  \label{fig:confusion matrix}
\end{figure}

\noindent Our model is tasked with categorizing text into one of three labels: non-offensive, direct violence, and passive violence. The confusion matrix, displayed in Figure~\ref{fig:confusion matrix}, depicts the performance of the model across these categories. It's pivotal to recognize that in our task, an ideal model would demonstrate high precision and recall across all three labels. 

\noindent The model categorizes text into non-violence (label 0), passive violence (label 1), and direct violence (label 2) with an overall macro F1 score of 0.74. It particularly excels in identifying non-violence texts. It also demonstrates aptitude in recognizing passive violence texts. However, it faces challenges in the realm of direct violence.

\section{Conclusion}

In this paper we described the nlpBDpatriots approach to the VITD shared task. We evaluated various models on the data provided by the shared task organizers, namely statistical machine learning models, transformer-based models, few shot prompting, and some customization with transformer-based models with multilinguality, back translation, and two-step classification. We show that the two-step classification procedure with multilinguality and back translation is the most successful approach achieving a macro F1 score of 0.74. 

Our two-step approach towards solving the problem presented for this shared task shows promising results. However, the relatively small size of the dataset made it difficult for the other pre-trained models to learn informative features that would help them perform classification. Also, the dataset contains three imbalanced labels making it easy for the models to overfit. 
Our approach with data augmentation and two-step classification generates good results, but it is still below one of the three baseline results announced by the organizers prior to the start of the competition. 

\section*{Acknowledgment}

We would like to thank the VITD shared task organizing for proposing this interesting shared task. We further thank the anonymous reviewers for their valuable feedback. 

\bibliography{anthology,custom}
\bibliographystyle{acl_natbib}






\end{document}